\documentclass[french]{pfia}
\usepackage{hyperref}
\usepackage[T1]{fontenc}
\usepackage{xcolor}

\usepackage{graphicx}
\newcommand{\Sarbacanes}{{\sc Sarbacanes~}}

\newcommand{\Ardans}{Ardans~}
\newcommand{\TAKM}{{\sc TA~KM~}}
\newcommand{\TA}{TechnicAtome~}
\newcommand{\akm}{Ardans Knowledge Maker\up{\textregistered} }  
\newcommand{\make}{Ardans {\it make}\up{\textregistered} }  
\newcommand{\sic}[1]{\og{}{\it #1 }\fg{}}
\newcommand{\iso}{{\sc ISO30401}}
\newcommand{\guil}[1]{%
 \ifthenelse{\isempty{#1}}%
    {}
    {\og{}#1\fg{}}
}
\title{\textbf{Implanter une approche hybride dans une démarche d’ingénierie de la connaissance pour manager les avis techniques relatifs au retour d'expérience d'exploitation d'un équipement sensible complexe}}

\author{S. Boblet\fup{1}, T. Cartié\fup{1}, A. Berger\fup{2}, J-P. Cotton\fup{2}, F. Vexler\fup{2} \\[6pt]
\fup{1} TechnicAtome\\  Lieu-dit « Les Hautes Rives », Route de Saint Aubin\\ 91190 Villiers-le-Bâcle, France - \url{www.technicatome.com}\\ \{sebastien.boblet, thierry.cartie\}@technicatome.com\\
\fup{2} Ardans SAS\\		6 rue Jean Pierre Timbaud, \og{}Le Campus\fg{ } Bâtiment B1\\78180 Montigny-le-Bretonneux, France - \url{www.ardans.fr}\\ \{aberger, jpcotton, fvexler\}@ardans.fr\\
}

\begin{document}

\maketitle


\begin{resume}
Comment manager les avis techniques relatifs au retour d'expérience d'exploitation de manière efficiente dans une organisation qui n'a jamais fait appel aux techniques et méthode de l’ingénierie de la connaissance?
Cet article précise comment un industriel du nucléaire et du secteur de la défense s'est approprié une telle démarche adaptée à son contexte organisationnel \sic{TA KM} qui s'inscrit dans le cadre de l'\iso, pour construire un système complet avec une application \sic{\Sarbacanes} pour supporter son process métier et pour pérenniser son savoir-faire et ses expertises dans une base de connaissance.
Au-delà du très classique transfert de connaissance entre expert et spécialiste métier, \Sarbacanes révèle aussi la capacité d’une telle ingénierie à offrir comme résultat une exploitation polyfonctionnelle. La modélisation a été accélérée par la mise en œuvre d’un outil adapté à ce type d’opération : la plate-forme \akm.
\end{resume}

\begin{motscles}
\Sarbacanes, Ingénierie de la connaissance, Gestion de Retour d'expérience, Modélisation de process métier, Recueil d’expertise, Gestion et Management, Sûreté nucléaire, Exploitation documentaire, Transfert de connaissance, \TAKM, \akm, \iso.
\end{motscles}

\begin{abstract}
How can technical advice on operating experience feedback be managed efficiently in an organization that has never used knowledge engineering techniques and methods? This article explains how an industrial company in the nuclear and defense sectors adopted such an approach, adapted to its \sic{\TAKM} organizational context and falls within the \iso ~framework, to build a complete system with a \sic{\Sarbacanes} application to support its business processes and perpetuate its know-how and expertise in a knowledge base.
know-how and expertise in a knowledge base. Over and above the classic transfer of knowledge between experts and business specialists, \Sarbacanes also reveals the ability of this type of engineering to deliver multi-functional operation. Modeling was accelerated by the use of a tool adapted to this type of operation : the \akm platform
\end{abstract}



\section{Introduction}
Depuis 2019, les sociétés \TA et \Ardans travaillent sur la mise en place d'un \sic{système métier} qui allie méthode d'ingénierie de la connaissance, plate-forme de gestion de la connaissance, intégration au système d'information de production, conduite de changement pour un meilleur service rendu au client final fondé sur la meilleure efficience opérationnelle des spécialistes et experts de la société (cf. Cartié \& al.~\cite{FIIA21TA}).
Cet article intègre la vision opérationnelle, la méthode d'élaboration du dispositif, la nature de l'approche hybride retenue (cf. Boblet \& al. ~\cite{TAKMHyCHA2024})  et le retour d'expérience de la mise en exploitation et de la dynamique qui en découlent.

\section{La description du contexte opérationnel }
L'acteur opérationnel est un industriel du secteur de la défense, la société \TA . Le domaine concerné est la gestion du retour d'expérience (REX) (cf. Malvache \& al. ~\cite{Malvache-93}) d'exploitation des équipements techniques que \TA conçoit et livre aux forces.

\begin{figure}[ht]
    \centering
    \includegraphics[width=9cm]{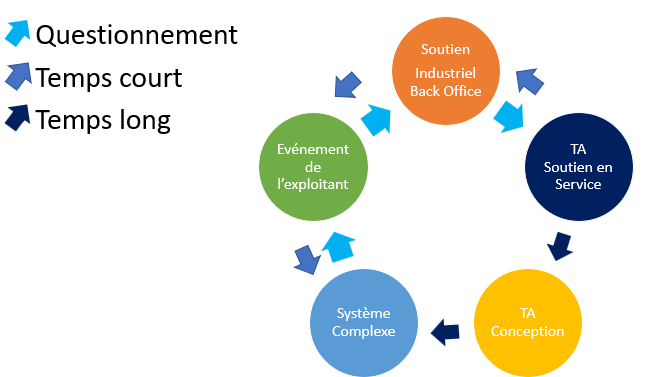}
    \caption{Le retour d'expérience}
    \label{Fig_SarbacanesTAKM_REX}
\end{figure}

Le sujet est le suivant : à partir d'un événement en exploitation sur l'équipement qui est un système complexe, la caractérisation du fait technique remonte à deux niveaux : soutien industriel puis soutien en service. Ainsi que le représente la Figure~\ref{Fig_SarbacanesTAKM_REX}, les experts doivent retourner dans un \sic{temps court} à l'exploitant un avis technique.
Par ailleurs, dans un \sic{temps long }, cette analyse est complétée avec les concepteurs pour éventuellement gérer une intervention voire intégrer des évolutions lors d'une opération de maintenance future. Il s'agit de considérer alors l'impact sur le système de soutien ou bien la prise en compte des évolutions en conception dans le cas d'un impact sur le système principal.
Les objectifs opérationnels sont nombreux : depuis l'optimisation des processus de capitalisation de REX jusqu'à l'analyse proprement dite du REX. Il y a aussi une priorité sur l'efficience dans l'analyse de l'événement en exploitation par rapport à tous les événements déjà survenus et analysés. Il y a bien sûr le fait de disposer d'un meilleur accès à la connaissance collective, et une fluidité dans les échanges entre exploitants et mainteneurs, comme entre mainteneurs et concepteurs. Enfin, l'évolution culturelle est un enjeu sous-jacent, avec la mise en place d'une approche collaborative permanente pour obtenir une amélioration continue effective. 

{\it In fine}, l'objectif est de fournir la garantie de la qualité et de la fidélité du REX pour tous les acteurs impliqués.

\section{La méthode \TAKM pour conduire l'opération dans le temps}

{\it "Construire en commun un objet inconnu" } (cf. Grundstein ~\cite{Grundstein94} ) nécessite un processus de confiance pour agréger au fil de l'eau cette matière ”connaissance”. Avant de s'interroger sur la technologie informatique d'IA il y a la question de la méthode et du phasage de l'opération afin d'être dans un processus d'appropriation et d'adhésion de la démarche ”étape par étape”.

\begin{figure}[ht]
    \centering
    \includegraphics[width=9cm]{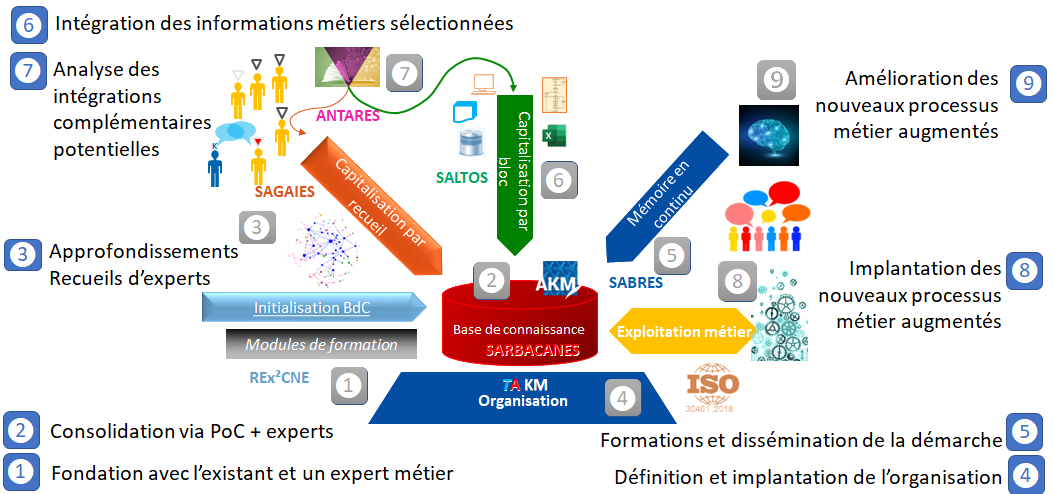}
    \caption{Les différentes étapes de la méthode \TAKM et ses facettes}
    \label{Fig_SarbacanesTAKM_Demarche}
\end{figure}

Initialisation de la modélisation, structuration de la connaissance, nourrissage des éléments de la base de connaissance, validation de ces mêmes contenus, enrichissement par de nouveaux éléments d'expertise, injection d'éléments de connaissance en masse, formation des utilisateurs pilotes, élicitation du processus métier complet et implantation dans l'environnement informatique de production : voilà en quelques activités le squelette de ce que porte la méthode \TAKM qui a conduit et irrigué l'opération depuis son lancement en 2019 (cf. Figure~\ref{Fig_SarbacanesTAKM_Demarche}) et selon les principes de l'\iso  ~(cf. Berger~\cite{berger:hal-04152777}). On considère à ce stade que \TAKM porte le \sic{système de gestion de la connaissance} sur le métier, ce qui s'entend comme le {\it KMS} (Le KMS pour \sic{Knowledge Management System} intègre la plate-forme logicielle à l'organisation de la gouvernance de la connaissance) au sens de l'\iso ~\cite{iso30401}.

\section{Le positionnement de l'approche hybride}
L'approche hybride s'appuie sur une méthode d'ingénierie des connaissances outillée éprouvée : \make et \akm. 
Afin d'apprécier \akm, la méthode \make et l'approche hybride {\it symbolique \&  sémantico syntaxique} le lecteur est invité à parcourir les articles ayant déjà traité de ces sujets à savoir : Mariot \& al. ~\cite{EGC2007RNTI}, Besson \& al. ~\cite{AKMEGC15}, Berger ~\cite{AKMAPIA2015}, Vexler \& al.~\cite{AKMEGC2020}, Mary \& al.~\cite{AKMAPIA2020} et Fourtout \& al.~\cite{EpioneAPIA2023}.

L'hybridation est relative d'une part à une {\bf modélisation symbolique} où des éléments de connaissances (on parle aussi d'articles ou de fiches) sont produits à partir de modèles, reliés à une ontologie élaborée au fil de la démarche, complétée par les liages inter-fiches validés par les sachants et une gestion de droits d'accès, et d'autre part à une {\bf approche sémantico syntaxique} réalisé par apprentissage incrémental sur les contenus des éléments présents dans la base de connaissance et enrichi lors de la complétion d'une nouvelle fiche.

\begin{figure}[ht]
    \centering
    \includegraphics[width=9cm]{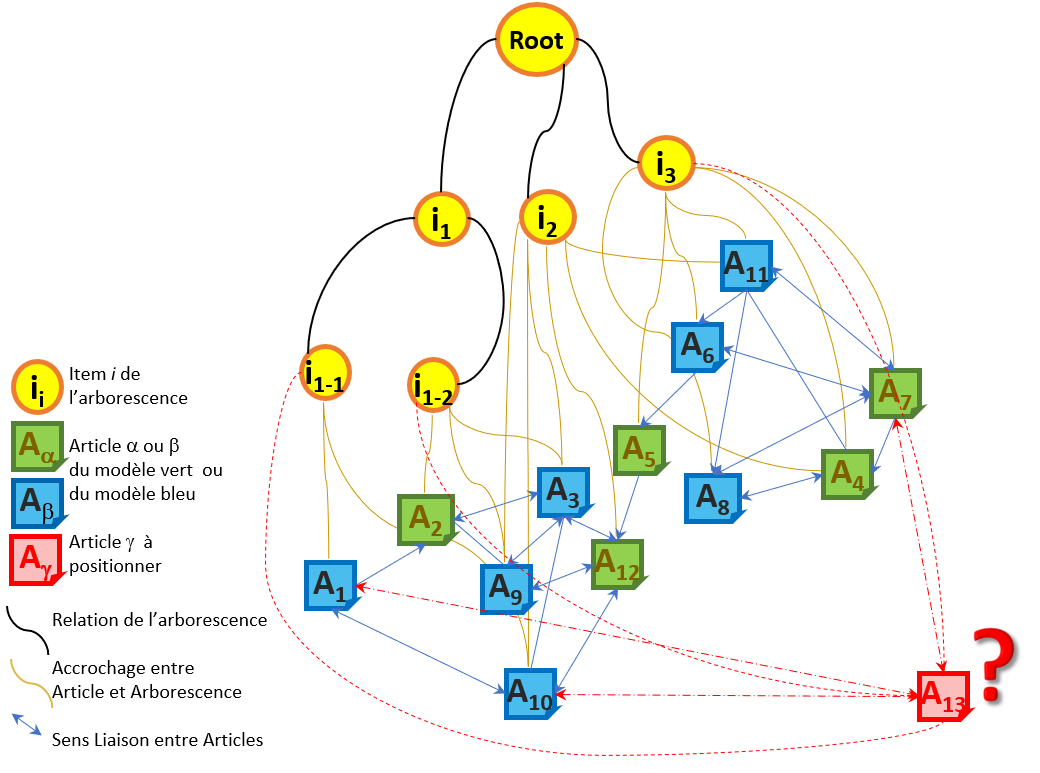}
    \caption{La puissance de l’approche hybride : la modélisation associée à une approche sémantico syntaxique}
    \label{Fig_SarbacanesTAKM_AKM}
\end{figure}

Le côté hybride a été nécessaire afin de satisfaire une ergonomie cognitive pour deux fonctionnalités essentielles (cf. le point d'interrogation de l'article A{\scriptsize 13} avec ses deux types de liens vers les items de l'ontologie ou vers les autres   Figure~\ref{Fig_SarbacanesTAKM_AKM}).

{\bf La première est : } comment être sûr de trouver les événements les plus proches de celui qu'il convient d'étudier et qui ont déjà été analysés et expertisés par le passé ?

{\bf La seconde est : } comment garantir que l'élément de connaissance qui va être créé, est positionné correctement avec le liage pertinent (La \sic{pertinence} est une conséquence de l'application de la méthode qui impose que les liens entre les éléments posés soient validés par les experts) au sein du réseau qui contient plusieurs dizaine de milliers d'éléments ?

\section{Le retour d'expérience après la mise en service et la première phase de déploiement }
\subsection{Le calendrier de \Sarbacanes}
\begin{figure}[ht]
    \centering
    \includegraphics[width=9cm]{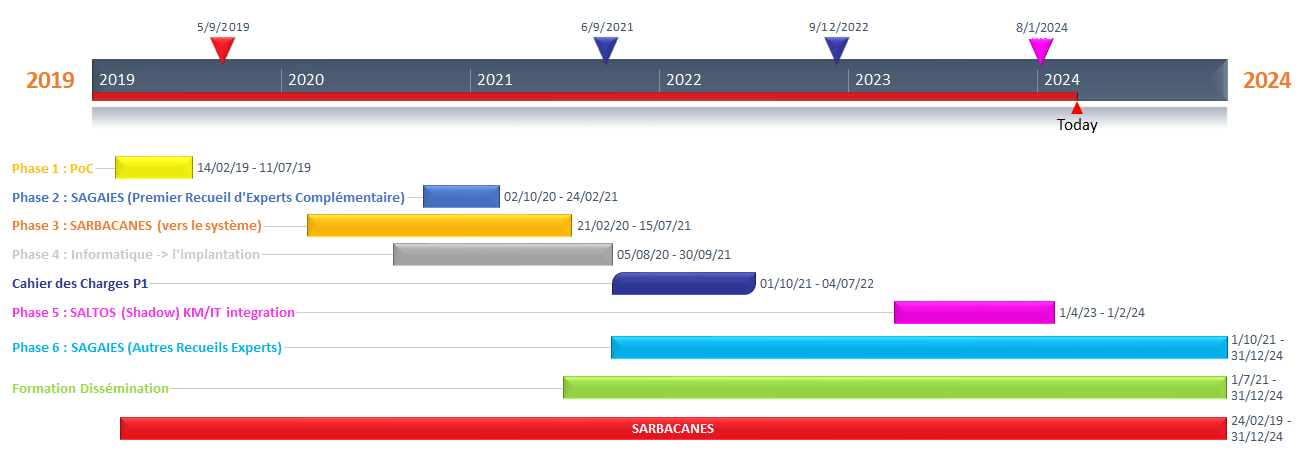}
    \caption{Les éléments majeurs du calendrier de l’opération industrielle}
    \label{Fig_SarbacanesTAKM_Agenda}
\end{figure}	

La conduite de ce type d'opération de management de REX (cf. Figure~\ref{Fig_SarbacanesTAKM_Agenda}) soulève des aspects de complexité initialement cachés.
La réalité temporelle d'un projet basé sur de l'ingénierie de la connaissance et celle d'une appropriation à l'aune de la taille de l'organisation.
La question de la confiance avec un système d'IA hybride s'est ici construite incrémentalement ainsi que celle de l'appropriation quasi immédiate des acteurs métiers adhérents à la démarche.
La réussite de ce projet est aussi une implication forte de sponsor métier expérimenté et d'un moral à tout épreuve.
La satisfaction enfin de constater qu'après la mise en exploitation, de nouvelles fonctionnalités "métier" connexes sont demandées puis intégrées à la base de connaissance initiale pour être exploitées très naturellement dans la vie courante.

\subsection{Des éléments de modélisation métier de \Sarbacanes}

Voici la méta-modélisation sous-jacente à la base de connaissance \Sarbacanes.
\begin{figure}[ht]
    \centering
    \includegraphics[width=9cm]{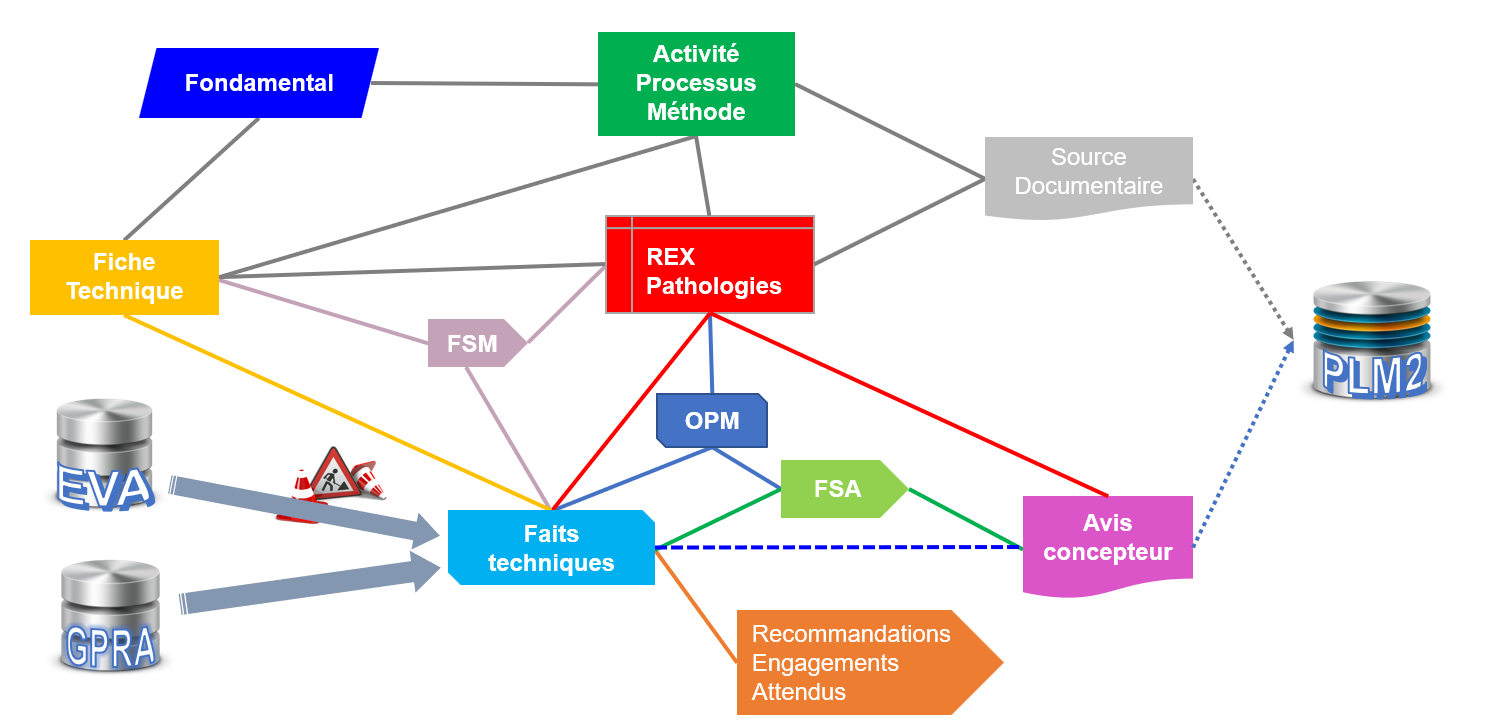}
    \caption{La structuration des modèles et les différentes étapes de son déploiement}
    \label{Fig_SarbacanesTAKM_Modeles}
\end{figure}	

Si l'on prend en compte les premiers types de modèles pour analyser un \sic{Fait Technique}, quelles sont les connaissances manipulées ?

\begin{center}
    \begin{tabular}{|p{2,8cm}|p{4,8cm}|} \hline
   Nom du Modèle  & Description : Exemple  \\ \hline\hline
  {\color{blue} Fondamental  } & Un principe ou phénomène : \par Ex. Corrosion des métaux \\ \hline  
  {\color{green} Activité Processus  } & Une action métier : \par Ex. {\'E}volution du plan de maintenance\\ \hline  
  {\color{orange} Fiche Technique}& Un équipement ou une solution technique : \par Ex. Architecture du contrôle-commande\\ \hline  
  {\color{gray}  Source \par Documentaire } & Une référence documentaire PLM : \par Ex. Procédure TA-6253301A  \\\hline  
  {\color{red}  REX Pathologie } & L'état de l'art sur un sujet : \par Ex. Phénomènes vibratoires \\ \hline  
  {\color{cyan} Fait Technique } & Un événement d'exploitation : \par Ex. Alarme sur circuit AUGM24 \\ \hline  
  {\color{magenta}  Avis Concepteur } & Diagnostic \& Prescription sur une situation : \par Ex. Possibilité de déroger à AB.SB.TC02 pour l'IPER \\ \hline  
   \hline
\end{tabular}
\end{center}

Par exemple, lorsqu'un \sic{{\color{cyan} Fait Technique }} est annoncé, il concerne un équipement renseigné par une \sic{{\color{orange} Fiche Technique }} au cours d'une \sic{{\color{green} Activité (Processus)  }} le tout étant conçu sur la base de \sic{{\color{blue} Fondamentaux }}. Les événements déjà instruits sur un sujet de même nature ont fait l'objet d'\sic{{\color{magenta}  Avis Concepteur(s) }} et les analyses antérieures ont été consolidées dans des \sic{{\color{red}  REX Pathologie }}  référencé(s) comme \sic{ {\color{gray}  Source documentaire }} dans des documents stockés dans l'outil de \sic{PLM}.

Si la réponse n'est pas forcément la solution, avec \Sarbacanes les ingénieurs en charge de l'analyse du nouveau \sic{Fait Technique}, savent ainsi immédiatement \sic{positionner} cette situation dans le patrimoine de \TA et débutent leur instruction en toute connaissance de cause.

\subsection{La vie de \Sarbacanes}
La connaissance s'agrège, se sédimente, s'incrémente au fil de l'eau et des retours d'expérience ou travaux nouveaux au plus grand plaisir des acteurs qui au quotidien voient leur travail prendre de la hauteur et un intérêt croissant.

Cette opération transcende l'équation de Davenport \& Pruzak ~\cite{Davenport98} car ici on observe que :
\begin{center}
    {\bf Knowledge Transfer = \\Transmission + Absorption \& Use \underline{+ Enrichment}}
\end{center}
Il s'avère que le retour d'exploitation de \Sarbacanes confirme le fait que le système a non seulement un intérêt dans l'usage attendu dans la {\bf production d'analyse temps court}, mais aussi dans le {\bf transfert de connaissance} pour l'accompagnement des utilisateurs nouveaux dans le métier à s'approprier l'antériorité des analyses déjà produites, et dans l'enrichissement par la consolidation de l'expertise dans la {\bf production d'analyse temps long}. Cette analyse temps long correspond à une réflexion {\it a posteriori} des acteurs : après la {\bf Transmission}, l'{\bf Appropriation} et l'{\bf Usage}, il y a un véritable {\bf Enrichissement} de la connaissance!

\section{Perspectives}

\begin{figure}[ht]
    \centering
    \includegraphics[width=9cm]{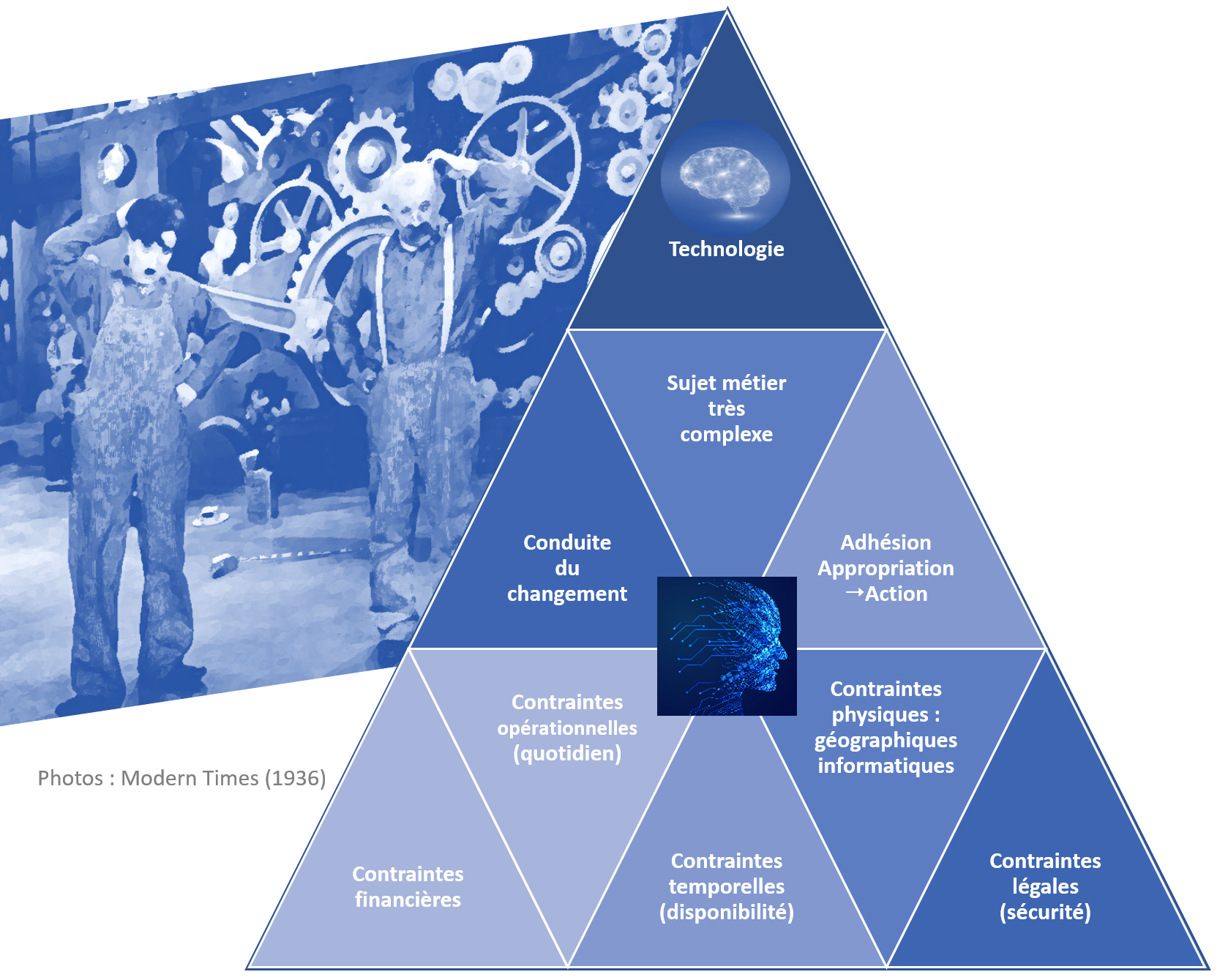}
    \caption{Où positionner la technologie, la connaissance, la confiance \& l’humain?  }
    \label{Fig_SarbacanesTAKM_Pyramide}
\end{figure}

Très concrètement, le projet se développe progressivement par les apports pragmatiques qui simplifient le travail quotidien des utilisateurs. La fait d'avoir accroché les connaissances au process métier et le lien aux sources qui justifient la  réponse à la question posée génère un cercle vertueux de confiance non pas simplement d'un utilisateur mais de toute l'équipe métier.
Résultat, des processus de gestion métier sont basculés et déployés au travers de \Sarbacanes ce qui fait évoluer les pratiques et ravi les acteurs qui disposent de plus de temps pour se concentrer sur le c\oe ur de leur métier. On observe enfin que la rigueur du processus de validation de cette connaissance collective relève d'une véritable \sic{hygiène d'ingénierie} pour l'organisation.

Une autre question est évoquée et est en cours d'analyse aujourd'hui : celle de l'apport effectif de l’IA générative (LLM+RAG) à ce stade. Faut-il risquer de perdre en qualité de performance et de confiance en introduisant un risque potentiel après avoir construit une base de connaissance robuste et riche forte de plusieurs dizaine de milliers d'actifs tangibles et validés humainement ?

Les derniers échanges avec la communauté HyCHA (cf. \url{https://hycha24.sciencesconf.org/}) ~\cite{TAKMHyCHA2024} confirment l'aspect délicat comme la complexité technique et humaine dans l’implantation opérationnelle (cf. Figure~\ref{Fig_SarbacanesTAKM_Pyramide}) d'une technologie IA hybride dans le milieu industriel. 

\section{Conclusion}
A la question du \sic{passage à l’échelle pour un système industriel, où positionner le point dur entre la technologie, la connaissance, la confiance \& l’humain ?} finalement la difficulté n'est pas celle que l'on attend!

L'approche hybride pour réaliser un tel système technique et opérationnel ne constitue pas {\it in fine} le point le plus difficile pour garantir le succès de ce type de projet ambitieux qui est d'abord une aventure humaine incroyable : ne dit-on pas que \sic{La prudence est mère de la sûreté}?

Ce qui reste certain, c'est que la démarche \TAKM a permis de construire la base \Sarbacanes et le système de gestion de la connaissance du maintien en condition en service de \TA est fondé sur la base d'une somme de petits succès. La dissémination de cette méthode d'ingénierie et de son ancrage dans le métier est un processus lent qui est en tous les cas particulièrement prometteur.

\section{Remerciements}
Nous remercions vivement \TA d'avoir autorisé cette communication qui démontre tout l’intérêt d’une approche outillée de l’ingénierie de la connaissance appliquée aux retours d'expérience dans les métiers de l’ingénierie de la maintenance (et plus d'ailleurs), qui traitent en particulier ceux relatifs à la sûreté et la sécurité nucléaire.

\bibliography{Biblio_Ardans_TAKM}

\end{document}